\documentclass{article}

\usepackage[numbers]{natbib}
\usepackage[preprint]{neurips_2024}

\usepackage[utf8]{inputenc} 
\usepackage[T1]{fontenc}    
\usepackage[dvipsnames]{xcolor}
\definecolor{cvprblue}{rgb}{0.21,0.49,0.74}
\usepackage[pagebackref,breaklinks,colorlinks,citecolor=cvprblue]{hyperref}
\usepackage{url}            
\usepackage{booktabs}       
\usepackage{amsfonts}       
\usepackage{nicefrac}       
\usepackage{microtype}      
\usepackage{graphicx} 
\usepackage{enumitem}
\usepackage{amsmath} 
\usepackage{subcaption}

\title{PivotMesh: Generic 3D Mesh Generation via \\ Pivot Vertices Guidance}

\author{%
Haohan Weng~$^1$\thanks{Work done during the internship at ShengShu\qquad$^\dagger$Corresponding author}~\quad
Yikai Wang~$^{2\dagger}$~\quad 
Tong Zhang~$^1$~\quad
C. L. Philip Chen~$^1$~\quad
Jun Zhu~$^{23}$
\\
$^1$~South China University of Technology\quad
$^2$~Tsinghua University\quad
$^3$~ShengShu
\\
\url{https://whaohan.github.io/pivotmesh}
}

\begin{document}

\maketitle

\begin{abstract}
  Generating compact and sharply detailed 3D meshes poses a significant challenge for current 3D generative models. Different from extracting dense meshes from neural representation, some recent works try to model the native mesh distribution (i.e., a set of triangles), which generates more compact results as humans crafted. However, due to the complexity and variety of mesh topology, these methods are typically limited to small datasets with specific categories and are hard to extend. In this paper, we introduce a generic and scalable mesh generation framework PivotMesh, which makes an initial attempt to extend the native mesh generation to large-scale datasets. 
  We employ a transformer-based auto-encoder to encode meshes into discrete tokens and decode them from face level to vertex level hierarchically. Subsequently, to model the complex typology, we first learn to generate pivot vertices as coarse mesh representation and then generate the complete mesh tokens with the same auto-regressive Transformer. This reduces the difficulty compared with directly modeling the mesh distribution and further improves the model controllability. PivotMesh demonstrates its versatility by effectively learning from both small datasets like Shapenet, and large-scale datasets like Objaverse and Objaverse-xl. Extensive experiments indicate that PivotMesh can generate compact and sharp 3D meshes across various categories, highlighting its great potential for native mesh modeling.
\end{abstract}

\section{Introduction}

\begin{figure}
\centering
\includegraphics[width=\textwidth]{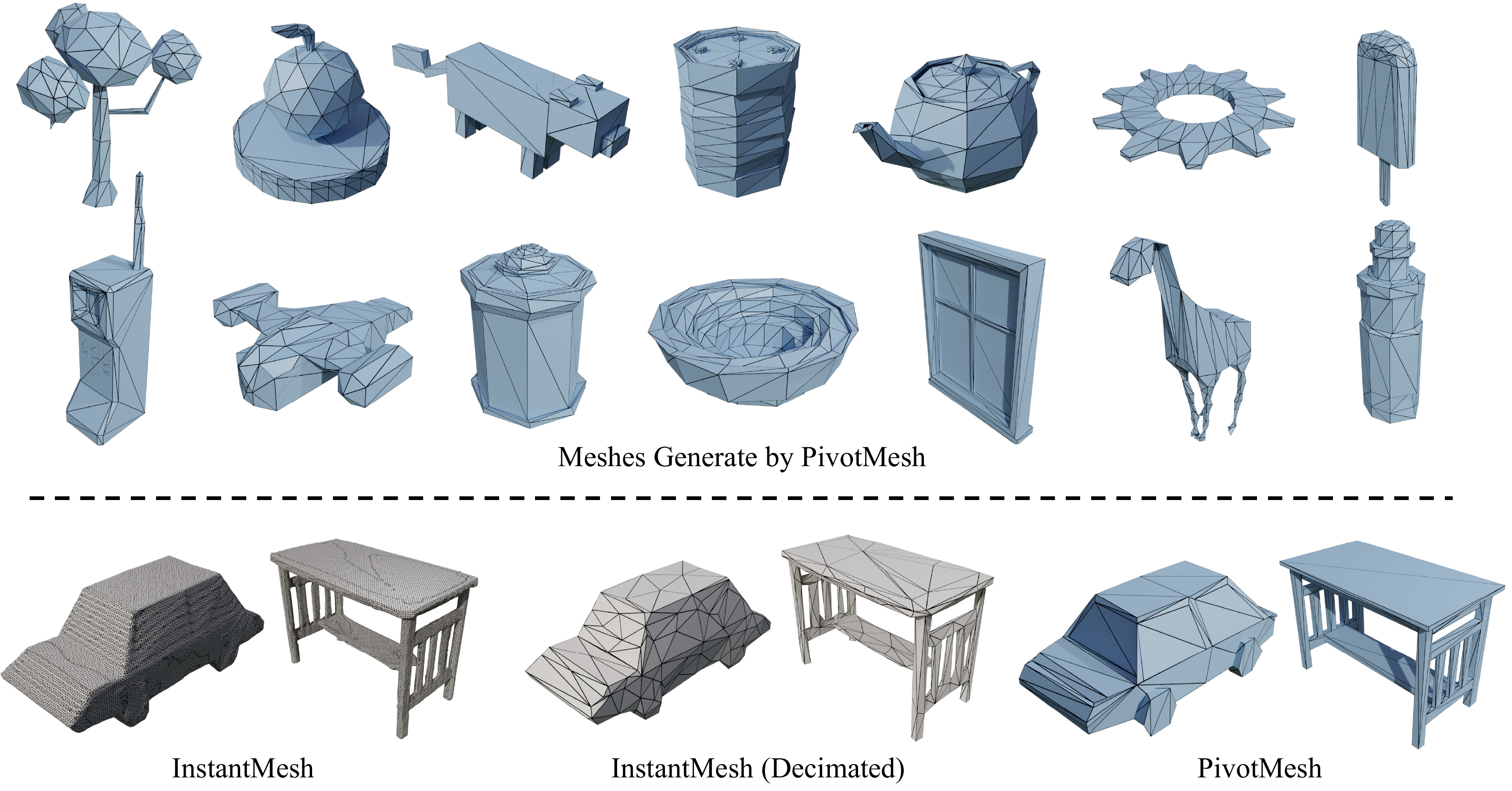}
\caption{
Different from 3D generation methods based on neural representations like InstantMesh \citep{xu2024instantmesh}, 
our methods can generate compact and sharp meshes with much fewer faces when producing similar shapes.
}
\vspace{-3mm}
\label{fig: head}
\end{figure}

The field of 3D generation has witnessed remarkable advancements in recent years \citep{pooleDreamFusionTextto3DUsing2023,hong2023lrm,xu2024instantmesh}. 
Meshes, the predominant representation for 3D geometry, are widely adopted across various applications from video games and movies to architectural modeling.
Despite the promising performance of current methods, they mostly rely on neural 3D representation like triplanes \citep{hong2023lrm,li2023instant3d} and FlexiCubes \citep{xu2024instantmesh}. 
Post-processed meshes extracted from these representations tend to be dense and over-smoothed, which are unfriendly for modern rendering pipelines as shown in Figure \ref{fig: head} (bottom). 
In contrast, meshes crafted by humans are typically more compact with fewer faces, reusing geometric primitives to efficiently represent real-world objects.

To avoid extracting dense meshes through post-processing, another promising direction is emerging that focuses on explicitly modeling the mesh distribution (i.e., native mesh generation).
This line of works \citep{nash2020polygen,siddiqui2023meshgpt,alliegro2023polydiff} generates meshes by predicting the 3D coordinates of faces, thus producing compact meshes as humans.
However, due to the complexity and variety of topological structures in meshes, these methods are typically confined to small datasets like Shapenet with single or narrow object categories, hindering the generalizability across diverse types of objects.
Therefore, \textit{it still remains a challenge to establish a generic generative model for native mesh generation at scale.}

In this paper, we propose PivotMesh, a generic and scalable framework to extend mesh generation to large-scale datasets across various categories.
PivotMesh consists of two parts: a mesh auto-encoder and a pivot-guided mesh generator. 
First, the auto-encoder is based on the Transformer to encode meshes into discrete tokens.
We also adopt a two-stage decoding strategy to decode mesh tokens from face level to vertex level hierarchically, which further improves the reconstruction performance and mesh surface continuity.
Second, we employ an auto-regressive Transformer to learn the joint distribution of pivot vertices and complete mesh tokens, where the pivot vertices serve as the coarse representation to guide the following mesh generation.
Specifically, pivot vertices are selected based on vertex degree and dropped randomly to prevent overfitting.
As shown in Figure \ref{fig: head} (top), once the model is trained, it can produce meshes from scratch, starting with the generation of pivot vertices followed by the complete mesh token sequence.
Furthermore, it can perform conditional generation given the pivot vertices from the reference mesh and supports downstream applications.

PivotMesh is designed to be scalable and extensible.
We initially evaluate its effectiveness on small dataset ShapeNet \citep{chang2015shapenet} as previous settings \citep{siddiqui2023meshgpt}.
Next, we carefully curate and train our model on the existing largest 3D datasets Objavese \citep{deitke2023objaverse} and Objaverse-xl \citep{deitke2024objaverse}. 
By leveraging large datasets, our model can generate generic meshes across various categories to accelerate the mesh creation process. 
Both the qualitative and quantitative experiments show that the proposed PivotMesh beats previous mesh generation methods like PolyGen \citep{nash2020polygen} and MeshGPT \citep{siddiqui2023meshgpt} by a large margin. 

The contributions of this paper can be summarized as follows:
\begin{itemize}[leftmargin=*]
    \item We propose a generic and scalable mesh generation framework PivotMesh, which makes an initial attempt to extend the native mesh generation to large-scale datasets. 
    \item We present a Transformer-based auto-encoder to preserve the geometry details and surface continuity in meshes by efficiently decoding from face level to vertex level hierarchically.
    \item We introduce pivot vertices guidance for complex mesh geometry modeling, which serves as the coarse representation to guide the complete mesh generation in a coarse-to-fine manner.
    \item PivotMesh achieves promising performance in various applications like mesh generation, variation, and refinement, accelerating the mesh creation process.
\end{itemize}

\section{Related Works}

\paragraph{Neural 3D Shape Generation.}
Most previous attempts learn 3D shape with various representations, e.g., SDF grids \citep{cheng2023sdfusion,chou2023diffusion,shim2023diffusion,zheng2023locally} and neural fields \citep{gupta3DGenTriplaneLatent2023,jun2023shap,muller2023diffrf,wang2023rodin,zhang20233dshape2vecset,liuMeshDiffusionScorebasedGenerative2023,lyu2023controllable}.
To improve the generalization ability, researchers start to leverage pretrained 2D diffusion models \citep{rombachHighResolutionImageSynthesis2022,saharia2022photorealistic,liuZero1to3ZeroshotOne2023} with score distillation loss \citep{pooleDreamFusionTextto3DUsing2023,linMagic3DHighResolutionTextto3D2023,wangProlificDreamerHighFidelityDiverse2023} in a per-shape optimization manner.
Multi-view diffusion models \citep{shi2023mvdream,weng2023consistent123,zheng2023free3d,shi2023zero123++,chen2024v3d,voleti2024sv3d} are used to further enhance the quality and alleviate the Janus problem.
Recently, Large Reconstruction Models (LRM) \citep{hong2023lrm,li2023instant3d,xu2023dmv3d,wang2024crm,xu2024grm,tang2024lgm,xu2024instantmesh} train the Transformer backbone on large scale dataset \citep{deitke2023objaverse} to effectively generates generic neural 3D representation and shows the great performance of scaling.
However, these neural 3D shape generation methods require post conversion \citep{lorensen1998marching,shen2021deep} for downstream applications, which is non-trivial and easy to produce dense and over-smooth meshes.

\paragraph{Native Mesh Generation.}

Compared with the well-developed generative models of neural shape representations, the generation of the mesh remains under-explored.
Some pioneering works try to tackle this problem by formulating the mesh representation as surface patches \citep{groueix2018papier}, deformed ellipsoid \citep{wang2018pixel2mesh}, mesh graph \citep{dai2019scan2mesh} and binary space partitioning \citep{chen2020bsp}.
PolyGen \citep{nash2020polygen} uses two separated auto-regressive Transformers to learn vertex and face distribution respectively. 
Polydiff \citep{alliegro2023polydiff} learns the triangle soups of mesh with a diffusion model.
MeshGPT \citep{siddiqui2023meshgpt} is most relevant to our work, which first tokenizes the mesh representation with a GNN-based encoder and learns the mesh tokens with a GPT-style Transformer. 
Despite its promising results on small datasets, the capability of MeshGPT is limited due to the design of its tokenizer and sequence formulation.
Furthermore, the performance of these works is constrained by the 3D data scale and model generalizability, constraining the downstream applications of mesh generation.
Unlike the above works, PivotMesh makes an initial attempt to build a generic generative model for native mesh generation within large-scale datasets.

\section{Method}

\begin{figure}
\centering
\includegraphics[width=\textwidth]{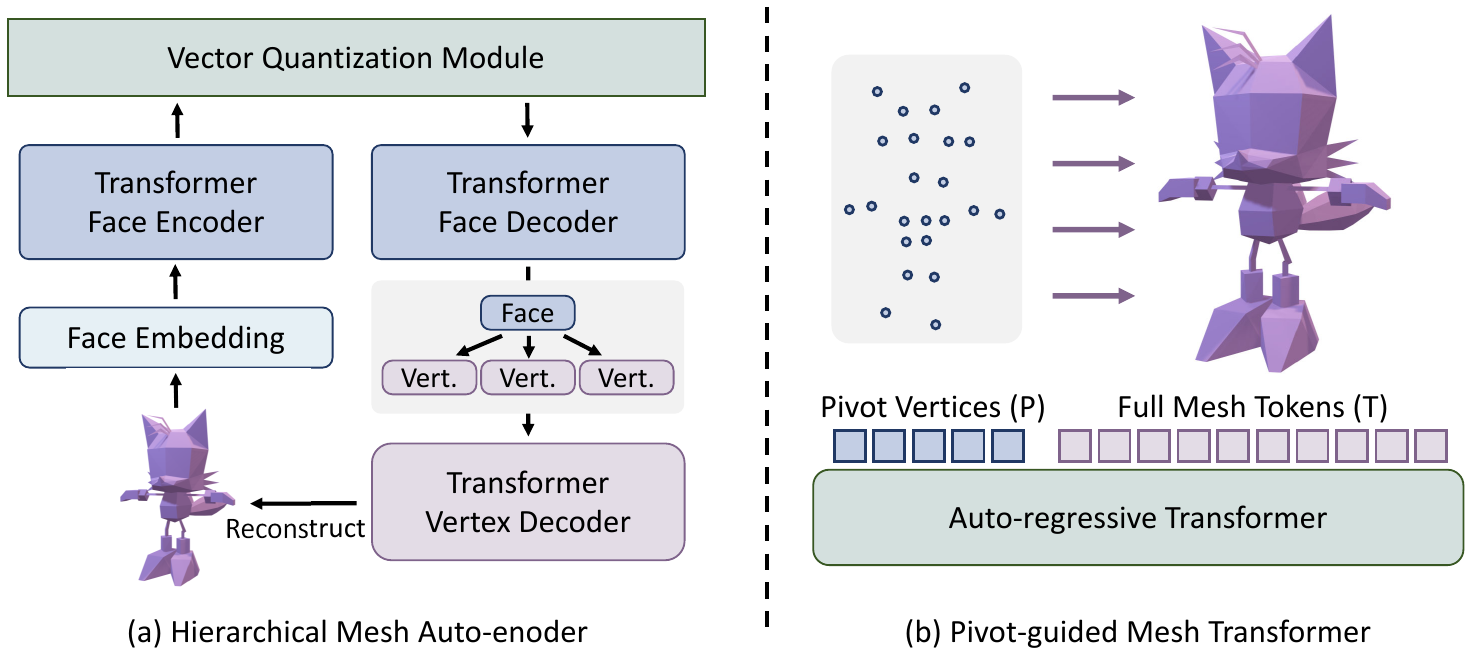}
\caption{\textbf{The overall method of PivotMesh.} (a) Triangle mesh sequences are tokenized into mesh tokens and hierarchically decoded from face level to vertex level via our mesh auto-encoder. (b) The auto-regressive Transformer first learns to generate pivot vertices as coarse mesh representation and then generates the complete mesh tokens in a coarse-to-fine manner.
}
\label{fig: method}
\end{figure}

In this section, we will introduce the details of the proposed PivotMesh as shown in Figure \ref{fig: method}. The challenges to scale up native mesh generation are analyzed in Section \ref{sec: method-challenges}.
Meshes formulated as triangle face sequences are first encoded into discrete tokens by the proposed mesh auto-encoder (Section \ref{sec: method-AE}). Then, we use an auto-regressive Transformer to learn the joint distribution of pivot vertices and mesh tokens (Section \ref{sec: method-pivotGPT}).

\subsection{Challenges for Native Mesh Generation on Large Datasets}
\label{sec: method-challenges}
There are two main challenges for scaling up native mesh generation to large datasets.

\textbf{Mesh Reconstruction.}
It is challenging to tokenize meshes into tokens due to the high requirement on reconstruction performance to preserve the mesh surface continuity.
Previous works like MeshGPT \citep{siddiqui2023meshgpt} formulate meshes as face graphs for reconstruction, which only focuses on face-level relationships and neglects the connection and interaction among vertices.
Furthermore, the limited network capability of auto-encoder (i.e., GNN and CNN) also hinders its scalability on large-scale datasets. 
To this end, we propose a Transformer-based auto-encoder to preserve the geometry details and surface continuity by decoding from face level to vertex level hierarchically.

\textbf{Complex Typology Modeling.}
Due to the complexity and variety of mesh topology, directly modeling the mesh sequence on large-scale datasets makes it easy to produce trivial meshes with simple geometry like cubes. 
It remains challenging to model complex typology in mesh structure.
For this purpose, we propose to first model a coarse representation of meshes (i.e., pivot vertices), then generate the full meshes in a coarse-to-fine manner.

\subsection{Encode Meshes into Discrete Tokens}
\label{sec: method-AE}

A triangle mesh $\mathcal{M}$ with $n$ faces can be formulated as the following sequence:
\begin{equation}
    \mathcal{M} := (f_1, f_2, ..., f_n) = (v_{11}, v_{12}, v_{13}, v_{21}, v_{22}, v_{23}, ..., v_{n1}, v_{n2}, v_{n3}),
\end{equation}
where each face $f_i$ consists of 3 vertices and each vertex $v_i$ contains 3D coordinates  discretized with a 7-bit uniform quantization.
To effectively learn the mesh distribution, we first tokenize the sequence into discrete tokens with the proposed transformer-based Auto-encoder.

\paragraph{Attention-based Tokenizer.} 

We use a Transformer-based architecture as the backbone for the encoder, capturing the long-range relationship between faces.
Furthermore, we replace the vanilla positional encoding in the Transformer with a single-layer GNN to capture the local topology of meshes. 
This preserves the permutation invariance of faces with higher scalability, yielding more effective and robust token representation for meshes.
Specifically, the features of each face $f_i$ (e.g., vertices embedding, normal embedding) are concatenated as one face embedding $F_{i}$ and sent to the Transformer-based face encoder: 
\begin{equation}
    F_i = \text{FaceEnc}(\text{FaceEmbed}(f_i)),
\end{equation}
In the vector quantization module, each face embedding $F_{i}$ is converted back to vertices embedding $V_{i1:i3}$, aggregated by shared vertex indices, and quantized with residual vector quantization \citep{martinez2014stacked}.
Each vertex embedding is quantized with $r$ codes, thus the total length of the token sequence $T$ is $|T| = 3nr$.
Then, the quantized vertex tokens are reorganized to face embedding $F'_{i}$ after the quantization module.
Similar to \cite{siddiqui2023meshgpt}, we empirically found that such vertex quantization is much easier for generative modeling due to the repeated appearance of vertex tokens. 

\paragraph{Hierarchical Decoding.}

To further improve the reconstruction performance and mesh surface contiguity, we design a hierarchical decoder from face level to vertex level.
The face embedding $F'_i$ from the vector quantization module is first passed to a face-level decoder. Then, the decoded face embedding is converted to vertex embedding $V'_i$ by a simple multi-layer perception (MLP).
The vertex embedding is then decoded by a vertex-level decoder, whose architecture is similar to the face decoder except that its input sequence is 3 times longer.
\begin{equation}
    \begin{aligned}
        F'_i &= \text{FaceDec}(F'_i), \\
        V'_i &= \text{VertexDec}(\text{MLP}_{n\rightarrow3n}(F'_i)),
    \end{aligned}
\end{equation}
The final decoded vertex embedding $V_i$ is then converted to the quantized 3D coordinate logits $\in (1, 2, ..., 2^7)$ for each axis ($x$, $y$ and $z$) and computes the cross entropy with the input mesh sequence.
Such hierarchical architecture allows the connection and interaction among both face and vertex level, thus improving the reconstruction accuracy and surface continuity.

\subsection{Guide Mesh Generation with Pivot Vertices}
\label{sec: method-pivotGPT}

\paragraph{Pivot Vertices Selection.}

The way pivot vertices are selected greatly affects the performance of the following mesh generation. 
For this purpose, we design a degree-based selection strategy for informative pivot vertices. Specifically, a mesh can be regarded as a graph, where each vertex $v_i$ represents a node and the connection between vertices represents the edges. Then, we compute the vertex degree  $deg(v_i)$ in mesh graphs and select the pivot vertices set $P$ with the top-degree vertices.
The size of the pivot vertices set is proportional to the number of vertices with a fixed ratio $\eta_{select}$.
Furthermore, to prevent overfitting in pivot-to-mesh modeling, we randomly drop some pivot vertices with the ratio $\eta_{drop}$ of all vertices for each training iteration.
In our experiments, the select ratio $\eta_{select} = 15\%$ and the dropping ratio $\eta_{drop} = 5\%$, yielding the final pivot vertex ratio $\eta = 10\%$.
The benefits of our pivot selecting strategy fall into two aspects.
First, it leverages frequently occurring vertices (with higher degree), enabling the Transformer to utilize these as conditional tokens mesh sequence generation efficiently. 
Second, it tends to preserve intricate mesh details, as regions with finer geometry typically necessitate more faces and thus larger vertex degrees.

\paragraph{Coarse-to-fine Mesh Modeling.} 

We employ a standard auto-regressive Transformer decoder to learn the joint distribution of the pivot vertex tokens $p_i \in P$ and the complete mesh tokens $t_i \in T$.
A learnable start and end token are used to identify the beginning and end of the token sequence, while a pad token is used to separate the pivot vertex tokens and mesh tokens.
The order of both pivot vertices tokens and full mesh tokens is sorted by $z$-$y$-$x$ coordinates from lowest to highest.
Different from the Transformer in Section \ref{sec: method-AE}, we add absolute positional encoding here to indicate the position in the token sequence.
The token sequences are modeled with a Transformer with parameter $\theta$ by maximizing the log probability:
\begin{equation}
    \prod_{i=1}^{|T|} p(t_{i} | t_{1:i-1}, P; \theta) \prod_{j=1}^{|P|} p(p_{j} | p_{j:j-1}; \theta),
\end{equation}
With such formulation, the auto-regressive Transformer first learns to generate pivot vertex tokens $P$ as coarse mesh representation and then generates the complete mesh tokens $T$ in a coarse-to-fine manner.
In such a coarse-to-fine manner, our model can effectively learn the complex mesh typology, which can be easily extended to large-scale datasets.
Once the model is trained, it can produce meshes from scratch, starting with the generation of pivot vertices followed by the full mesh sequence.
Furthermore, different from previous mesh generation methods \citep{nash2020polygen,siddiqui2023meshgpt}, our model can perform conditional generation given the pivot vertices from the reference mesh, and also supports downstream applications like mesh variation and refinement.

\section{Experiment}

\subsection{Experiment Settings}

\paragraph{Datasets.}

Our model is trained on various classes and scales datasets, including ShapeNetV2 \citep{chang2015shapenet}, Objaverse \citep{deitke2023objaverse}, Objaverse-xl \citep{deitke2024objaverse}.
For ShapeNet, follow previous settings \citep{nash2020polygen,siddiqui2023meshgpt}, we use 4 subsets (chair, table, bench, lamp) and filter the faces less than 800 after plannar decimation (with a fixed angle tolerance $\alpha=10^{\circ}$).
For Objaverse and Objaverse-xl, we apply the data curation and filter the objects whose faces are less than 500 \textit{without decimation} to preserve the mesh quality. 
The final dataset size after filtering of Shapenet, Objaverse, and Objaverse-xl is around 10k, 40k, and 400k respectively.
For each dataset, we split 1k samples for testing, and leave the rest as the training data.
For data augmentation, we use random scaling on each axis (from 0.95 to 1.05) and random shifts ($\pm0.01$) to enhance the data diversity.

\paragraph{Baselines.}

We benchmark our approach against leading mesh generation methods: \textbf{PolyGen} \citep{nash2020polygen}, which generates polygonal meshes by first generating vertices followed by faces conditioned on the vertices with two separate Transformer; \textbf{MeshGPT} \citep{siddiqui2023meshgpt}, which tokenizes the mesh sequence into mesh tokens with a GNN-ResNet based Auto-encoder and learns the mesh tokens directly with an Auto-regressive Transformer.

\paragraph{Metrics.}

Following the evaluation settings of previous mesh generation methods \citep{siddiqui2023meshgpt,alliegro2023polydiff}, 
we evaluate the reconstruction quality by two metrics, triangle accuracy and l2 distance. 
We use the following metrics for mesh quality assessment: Minimum Matching Distance (MMD), Coverage (COV), and 1-Nearest-Neighbor Accuracy (1-NNA). 
For MMD, lower is better; for COV, higher is better; for 1-NNA, 50\% is the optimal. 
We use a Chamfer Distance (CD) distance measure for computing these metrics on 1024-dim point clouds uniformly sampled from meshes.

\paragraph{Implementation Details.}

For the auto-encoder, the face encoder has 12 layers with a hidden size of 512, the face decoder has 6 layers with a hidden size of 512, and the vertex decoder has 6 layers with a hidden size of 256.
For vector quantization, the number of residual quantizers $r = 2$, and the codebook is dynamically updated by exponential moving averaging with codebook size 16384 and codebook dimension 256.
It is trained on an 8×A100-80GB machine for around 1 day with a batch size of 64 for each GPU.
For auto-regressive Transformer, it has 24 layers with a hidden size of 1024.
It is trained on an 8×A100-80GB machine for around 3 days with batch size 12 for each GPU.
The temperature used for sampling is set to 0.5 to balance the quality and diversity.
We use flash attention for all Transformer architecture and fp16 mixed precision to speed up the training process. 
We use AdamW \cite{loshchilov2017decoupled} as the optimizer with $\beta_1 = 0.9$ and $\beta_2=0.99$ with a learning rate of $10^{-4}$ for all the experiments.

\begin{figure}
\centering
\includegraphics[width=\textwidth]{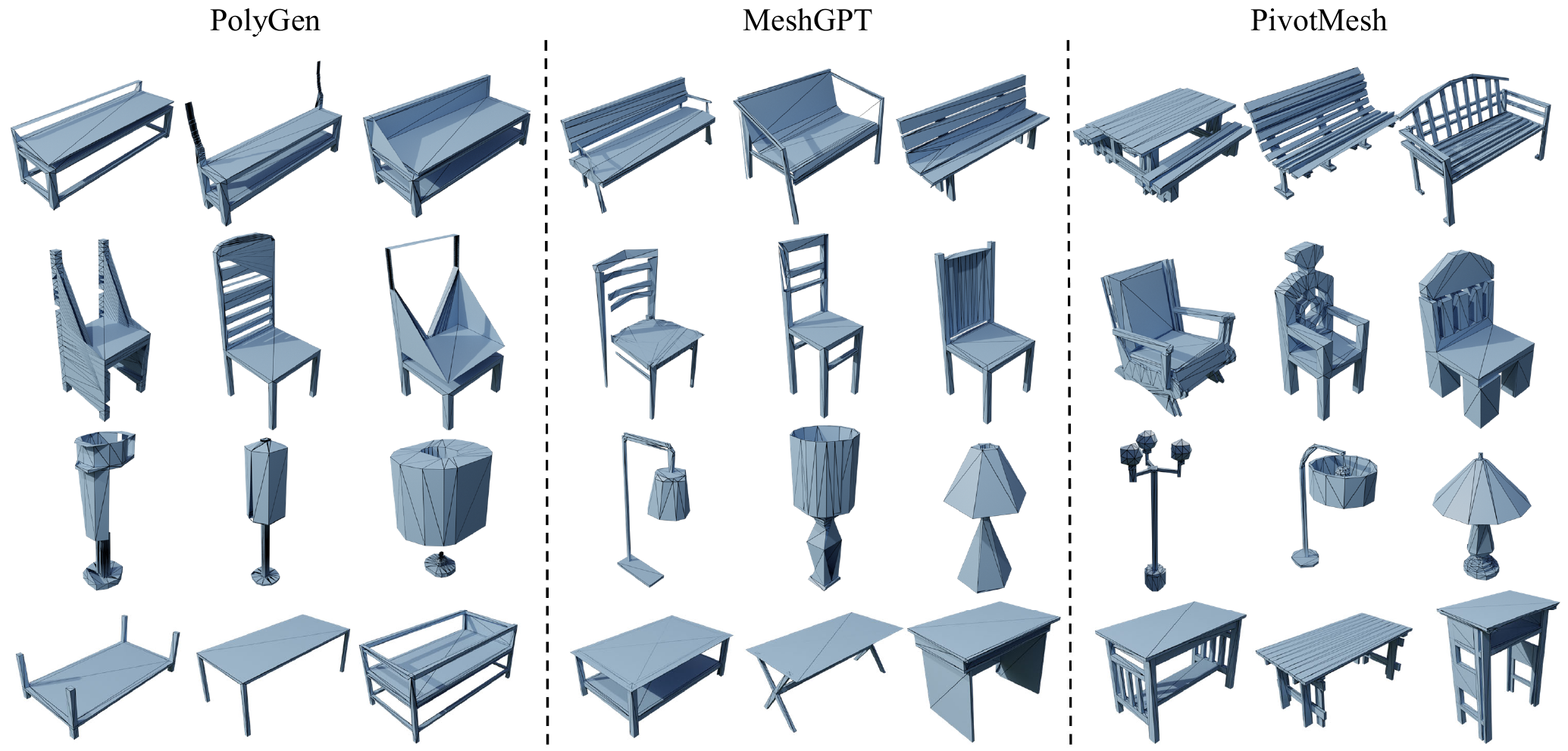}
\caption{\textbf{Qualitative comparison of unconditional generation on ShapeNet.} Each line represents a subset of ShapeNet (bench, chair, lamp, table). 
}
\label{fig: shapenet}
\end{figure}

\begin{table}[t]
\vspace{-3mm}
\caption{\textbf{The unconditional generation results on Shapenet dataset.} The proposed PivotMesh significantly improves the mesh quality compared with baselines by a large margin.}
\begin{center} 
\resizebox{\linewidth}{!}{
\begin{tabular}{c ccc ccc ccc} 
\toprule
\multicolumn{1}{r}{Model} & \multicolumn{3}{c}{PolyGen} & \multicolumn{3}{c}{MeshGPT}  & \multicolumn{3}{c}{PivotMesh}  \\
\cmidrule(lr){2-4} \cmidrule(lr){5-7} \cmidrule(lr){8-10} 
Subset
& COV(\%)$\uparrow$  & MMD($10^{-3}$)$\downarrow$ & 1-NNA(\%)$\downarrow$ 
& COV(\%)$\uparrow$  & MMD($10^{-3}$)$\downarrow$ & 1-NNA(\%)$\downarrow$ 
& COV(\%)$\uparrow$  & MMD($10^{-3}$)$\downarrow$ & 1-NNA(\%)$\downarrow$ \\
\midrule
Chair
&29.47	&13.34  &81.45
&41.31	&10.30  &61.84
&\textbf{52.89}	&\textbf{9.77}	&\textbf{56.71}
\\
Table
&38.67	&15.84   &66.27
&43.00	&9.77	&62.83
&\textbf{51.68}	&\textbf{9.28}	&\textbf{56.55}
\\
Bench
&37.50	&10.8	&79.69
&46.87	&9.48	&67.19
&\textbf{51.56}	&\textbf{8.50}	&\textbf{53.91}
\\
Lamp
&31.76	&33.87	&81.76
&45.88	&23.43	&57.06
&\textbf{50.58}	&\textbf{22.65}	&\textbf{54.71}
\\
\midrule
Mixed
&33.91	&13.47	&74.63
&43.89	&11.48	&63.24
&\textbf{50.42}	&\textbf{11.03}	&\textbf{60.89}
\\

\bottomrule \end{tabular} 
\vspace{-3mm}
}
\label{tab: shapenet}
\end{center} \end{table}

\subsection{Mesh Generation From Scratch}

\paragraph{Comparison with baselines on various-scale benchmarks.}

\begin{figure}
\centering
\includegraphics[width=\textwidth]{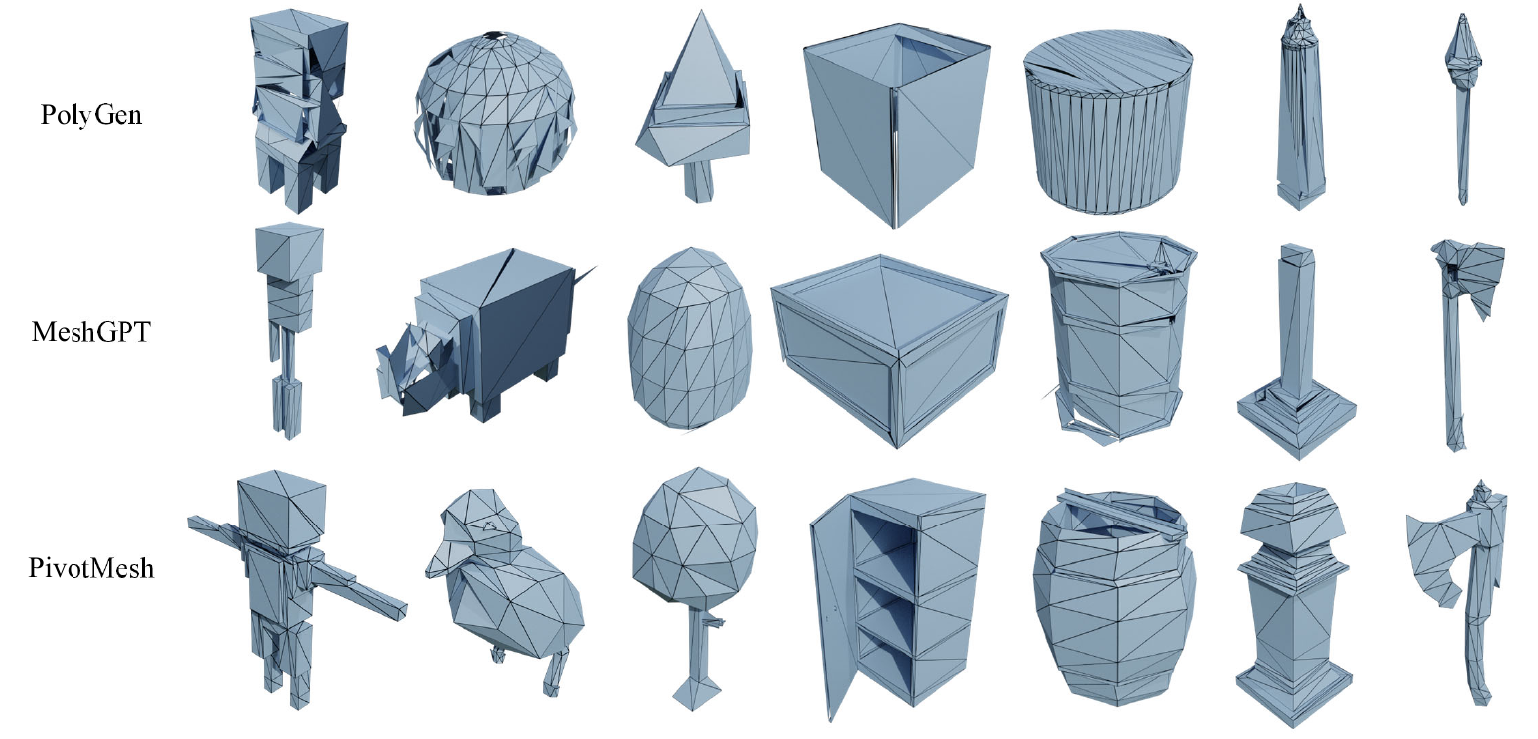}
\caption{\textbf{Qualitative comparison of unconditional generation on Objaverse.} 
}
\label{fig: objaverse}
\end{figure}

We first evaluate the proposed PivotMesh on the commonly used benchmark Shapenet with four selected categories, chair, table, bench, and lamp. Following the previous setting \citep{siddiqui2023meshgpt,alliegro2023polydiff}, we first pretrain our model on the mixture dataset of four selected categories and then finetune each category separately.
We report the generation results both on the mixed dataset and each subset in Table \ref{tab: shapenet}.
Furthermore, we train our model on the larger scale datasets Objaverse and Objaverse-xl and report the performance in Table \ref{tab: objaverse}.
For all these experiments, our method can achieve state-of-the-art performance on all evaluation metrics.
As shown in Figure \ref{fig: shapenet} and Figure \ref{fig: objaverse}, our model can generate meshes with the best visual quality and geometry complexity.
PolyGen often produces incomplete meshes due to the accumulation error by the separate training of the vertex model and face model.
MeshGPT can produce complete meshes but it is trapped in simple geometry due to its network capability and the complex mesh sequence.
With the hierarchical auto-encoder and pivot vertices guidance, our model can produce compact meshes with sharp details and complex geometry.

\begin{table}[t]
\vspace{-3mm}
\caption{\textbf{The unconditional generation results on the Objaverse and Objaverse-xl dataset.} 
}
\begin{center} 
\resizebox{0.8\linewidth}{!}{
\begin{tabular}{c ccc ccc} 
\toprule
\multicolumn{1}{r}{Model} & \multicolumn{3}{c}{Objaverse} & \multicolumn{3}{c}{Objaverse-xl} \\
\cmidrule(lr){2-4} \cmidrule(lr){5-7}
Dataset
& COV(\%)$\uparrow$  & MMD($10^{-3}$)$\downarrow$ & 1-NNA(\%)$\downarrow$ 
& COV(\%)$\uparrow$  & MMD($10^{-3}$)$\downarrow$ & 1-NNA(\%)$\downarrow$ \\
\midrule
PolyGen
&23.86	&24.01   &84.07
&21.79	&22.68   &83.40
\\
MeshGPT
&35.03	&17.30	&63.86
&41.50  &14.76  &64.25
\\
PivotMesh
&\textbf{46.48}	&\textbf{16.66}	&\textbf{58.55}
&\textbf{45.30}    &\textbf{14.33}     &\textbf{57.75}
\\
\bottomrule \end{tabular} 
}
\label{tab: objaverse}
\vspace{-3mm}
\end{center} \end{table}

\paragraph{Shape Novelty Analysis.}

\begin{figure}
\centering
\includegraphics[width=\textwidth]{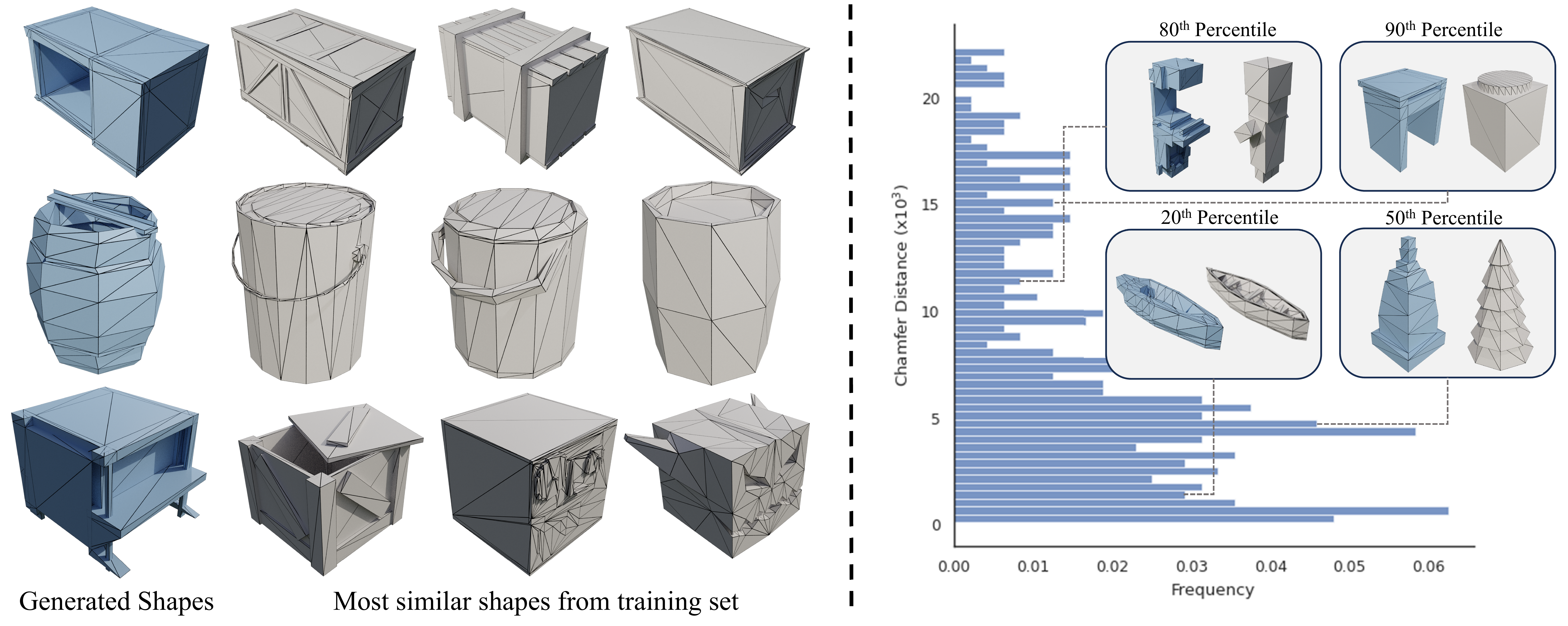}
\caption{\textbf{Shape novelty analysis on Objaverse dataset.} We show the 3 nearest neighbors measured in Chamfer Distance (CD) for generated shapes (left). We plot the distribution of 500 generated shapes from our method and their minimum CD to the training set (right). 
Shapes at the 50th percentile look different from the closest train shape.
It shows that our method not only covers shapes in the training set (low CD values) but also creates novel and realistic shapes (high CD values).
}
\label{fig: novolty}
\end{figure}

To show that our model can create novel shapes instead of memorizing the training set, we conduct a shape novelty analysis similar to \citep{hui2022neural,erkocc2023hyperdiffusion,siddiqui2023meshgpt}. 
We generate 500 shapes and search the 3 closest neighbors from the training set measured in Chamfer Distance (CD), shown in Figure \ref{fig: novolty} (left). We also show top-1 CD distribution between the generated shapes and the training set in Figure \ref{fig: novolty} (right).
The CD results show that our method not only covers shapes in the training set (low CD values) but also creates novel and realistic shapes (high CD values).

\begin{table}[t]
\vspace{-3mm}
\caption{\textbf{Ablation study of auto-encoder and auto-regressive Transformer on Objaverse.}}
\begin{subtable}[c]{0.47\textwidth}
\centering
\resizebox{\linewidth}{!}{
    \begin{tabular}{ccc}
    \toprule
    Method  & Triangle Accuracy(\%)$\uparrow$  & L2 Distance$\downarrow$ \\
    \midrule
    w/o Transformer   &86.89  &10.34 \\
    w/o hier. decode    &92.30    &5.37    \\
    Ours  &\textbf{97.89} &\textbf{1.05}         \\
    \bottomrule
    \end{tabular}}
\subcaption{Ablation Study on Auto-encoder.}
\end{subtable}
\begin{subtable}[c]{0.49\textwidth}
\centering
\resizebox{\linewidth}{!}{
    \begin{tabular}{cccc}
    \toprule
    Method  & COV(\%)$\uparrow$  & MMD($10^{-3}$)$\downarrow$ & 1-NNA(\%)$\downarrow$   \\
    \midrule
    w/o pivot guidance  &42.76 &16.83 &61.38 \\
    w/o pivot selection &42.48 &17.62 &62.90  \\
    Ours  &\textbf{46.48}	&\textbf{16.66}	&\textbf{58.55}  \\
    \bottomrule
    \end{tabular}}
\subcaption{Ablation Study on auto-regressive Transformer.}
\end{subtable}
\label{tab: ablation2}
\end{table}

\subsection{Ablation Study}

\paragraph{The effectiveness of Transformer-based hierarchical auto-encoder.} 

For mesh auto-encoder, previous works like MeshGPT \citep{siddiqui2023meshgpt} employ GNN as the encoder and CNN as the decoder. 
Table \ref{tab: ablation2} (w/o Transformer) shows the reconstruction performance of such architecture, which is significantly lower than our Transformer-based auto-encoder. 
The performance gain is mainly from the high capability and scalability of the Transformer compared with the GNNs and CNNs.
Table \ref{tab: ablation2} (w/o hier. decode) shows the performance of the auto-encoder when removing the hierarchical decoding mechanism.
It shows that such a hierarchical network design endows the decoder with better alignment in both face and vertex levels, thus improving the final reconstruction results.

\paragraph{The effectiveness of pivot vertices guidance and their selection.}

Table \ref{tab: ablation2} (w/o pivot guidance) shows the results of directly employing an auto-regressive Transformer to model the mesh tokens from our auto-encoder.
Without the pivot vertex guidance, the model fails to produce meshes with complex geometry.
In Table \ref{tab: ablation2} (w/o pivot selection), we employ random pivot vertex selection instead of the proposed degree-based strategy.
Some metrics from its results are even worse than that without using pivot vertices guidance.
The strong performance degradation is because the randomly selected vertices make it hard to summarize the whole geometry of the meshes and the Transformer is incapable of learning such pivot-mesh joint distribution, showing the importance of pivot vertices selection strategy.

\begin{figure}
\vspace{-3mm}
\centering
\includegraphics[width=0.9\textwidth]{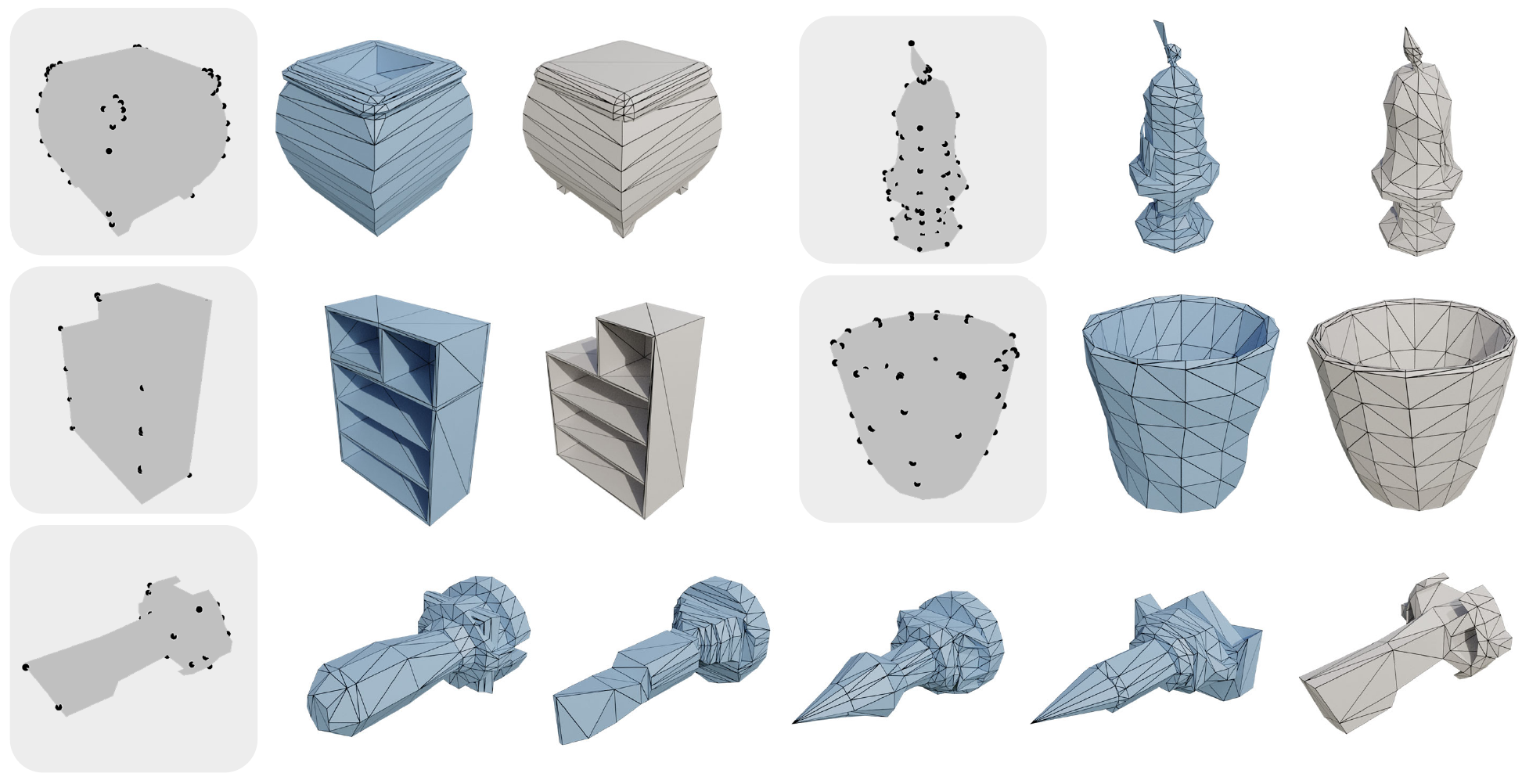}
\caption{\textbf{Conditional generation results with pivot vertices guidance from Objaverse test set.} The generated meshes are marked in blue and the ground truth meshes are marked in gray. The diversity of generation results is also shown on the bottom line.
}
\vspace{-3mm}
\label{fig: conditional}
\end{figure}

\vspace{-1mm}
\subsection{Pivot Vertices Guidance Analysis}
\vspace{-1mm}

To trade-off between the generalization and the visual quality for pivot-guided mesh generation, we first pretrain our model in the mixture of objaverse and objaverse-xl datasets. Then, we finetune the pretrained model on the well-curated objaverse to further improve the generated mesh quality.

\paragraph{Pivot-guided Mesh Generation.}

Given a reference mesh, we first encode it into mesh tokens and then select the pivot vertices. 
Our model can generate the corresponding meshes as shown in Figure \ref{fig: conditional}.
It shows that our model can generate diverse and high-quality meshes while maintaining the high-level structure corresponding to the pivot vertices.

\paragraph{Downstream Applications.}

\begin{figure}
\vspace{-3mm}
    \centering
    \begin{subfigure}{0.45\textwidth}
        \includegraphics[width=\textwidth]{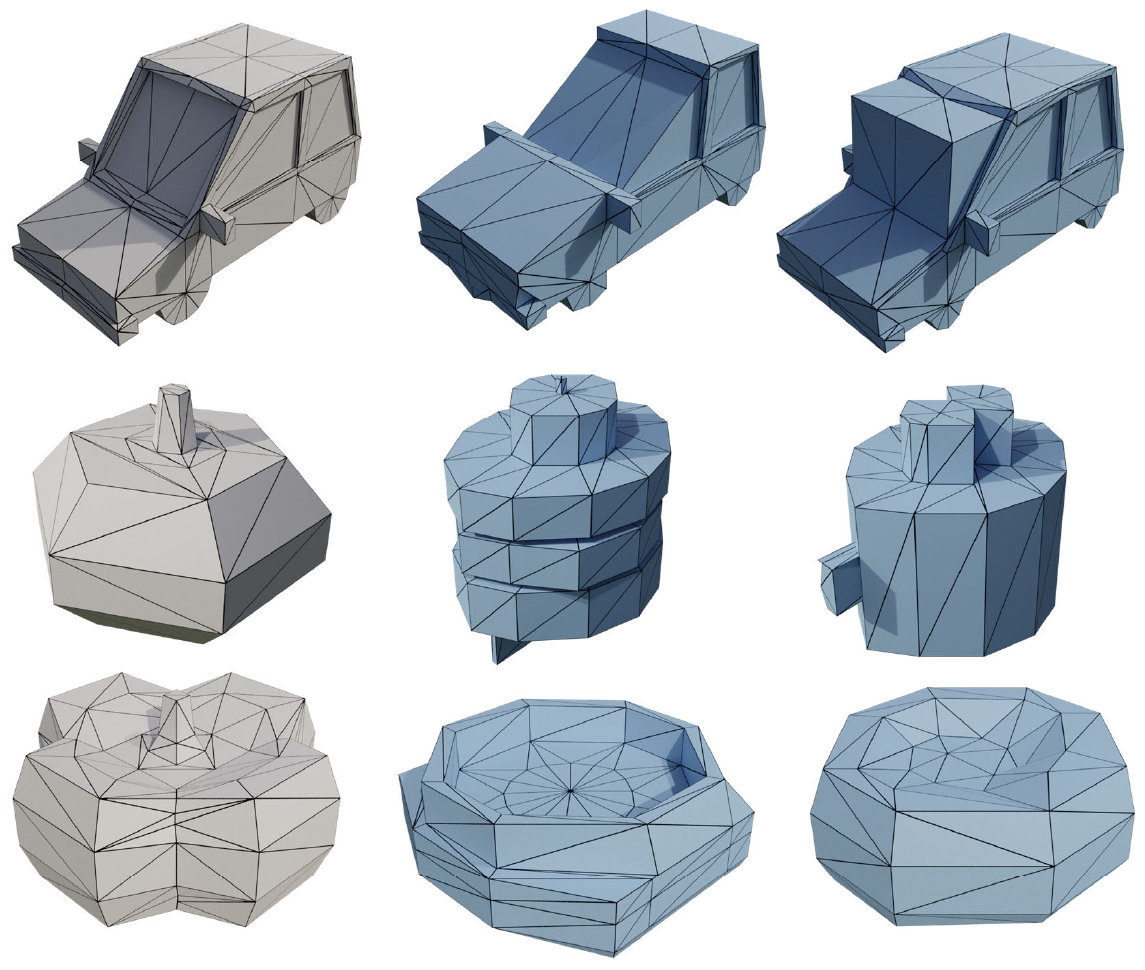} 
        \caption{Mesh Variation}
    \end{subfigure}
    \begin{subfigure}{0.45\textwidth}
        \includegraphics[width=\textwidth]{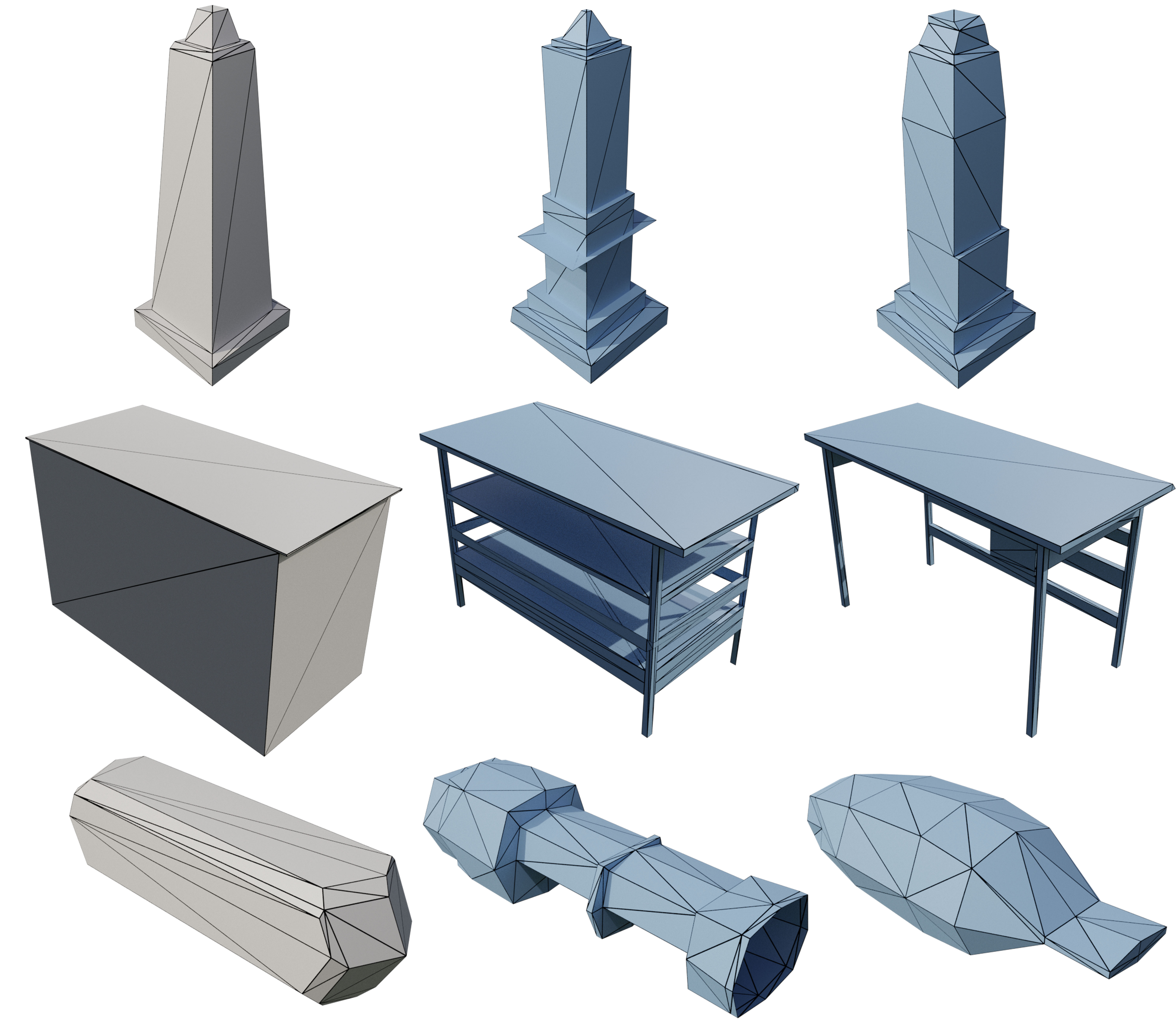} 
        \caption{Coarse Mesh Refinement}
    \end{subfigure}
    \caption{\textbf{PivotMesh can support various downstream applications} (Reference meshes are marked in gray). (a) Mesh variation: PivotMesh generates diverse meshes similar to reference meshes but with different details. (b) Coarse Mesh Refinement: PivotMesh can refine the details for coarse meshes to accelerate the mesh creation.
    }
    \label{fig: application}
\end{figure}

PivotMesh can serve as a generic mesh generative model to support various applications in Figure \ref{fig: application}. 
For mesh variation, our model can generate diverse variants with the user-given meshes.
For mesh refinement, it can be regarded as a special case of mesh variation, but with coarse meshes as input. Our model can refine the coarse meshes to fine meshes with detailed geometry to accelerate the mesh creation process.

\begin{figure}
\vspace{-3mm}
\centering
\includegraphics[width=\textwidth]{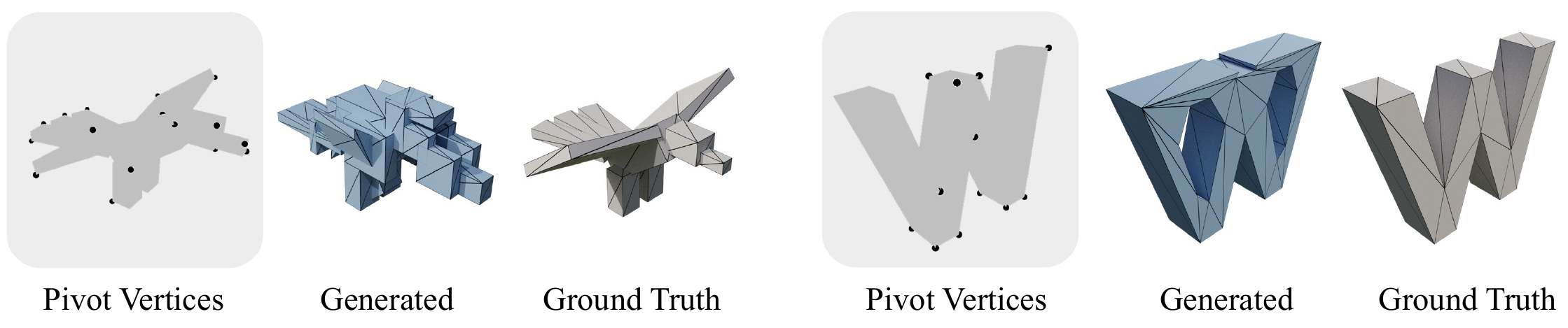}
\caption{\textbf{Some failure cases for the pivot-guided mesh generation.}
Since the same pivot vertices may indicate diverse meshes, our method may sometimes produce undesired geometry. This can be alleviated by adding more control conditions like images and texts.
}
\vspace{-4mm}
\label{fig: failure}
\end{figure}

\vspace{-2mm}
\subsection{Limitations and Future Work}

Although PivotMesh can produce compact meshes with high quality, it still has some limitations.
First, the controlling ability of PivotMesh is still not enough.
As shown in Figure \ref{fig: failure}, since the same pivot vertices may indicate diverse meshes, our method may sometimes produce undesired geometry. This can be alleviated by adding more control conditions like images and texts and we leave it for our future work.
Second, the scale of data and the number of model parameters is still limited.  
Due to the constrained computation resources, we only train PivotMesh on meshes less than 500 for both Objaverse and Objaverse-xl, and the number of model parameters is much less than recent advances in large language models \citep{Touvron2023Llama2O}. Given the increased data and model scale, the performance and capabilities of PivotMesh could be further boosted.

\section{Conclusion}

In this paper, we introduce PivotMesh, a scalable framework to generate generic meshes with compact and sharp geometry. By employing pivot vertices as a coarse representation to guide the mesh generation process and leveraging a Transformer-based hierarchical auto-encoder, PivotMesh demonstrates its capability to generate high-quality meshes on both small and large-scale datasets. 
The proposed model significantly outperforms existing methods and sets a new benchmark for native mesh generation tasks, showcasing its potential for creating novel shapes and supporting downstream applications.

\clearpage

{
\bibliographystyle{abbrv}
\bibliography{main}

\begin{thebibliography}{10}

\bibitem{alliegro2023polydiff}
A.~Alliegro, Y.~Siddiqui, T.~Tommasi, and M.~Nie{\ss}ner.
\newblock Polydiff: Generating 3d polygonal meshes with diffusion models.
\newblock {\em arXiv preprint arXiv:2312.11417}, 2023.

\bibitem{chang2015shapenet}
A.~X. Chang, T.~Funkhouser, L.~Guibas, P.~Hanrahan, Q.~Huang, Z.~Li, S.~Savarese, M.~Savva, S.~Song, H.~Su, et~al.
\newblock Shapenet: An information-rich 3d model repository.
\newblock {\em arXiv preprint arXiv:1512.03012}, 2015.

\bibitem{chen2020bsp}
Z.~Chen, A.~Tagliasacchi, and H.~Zhang.
\newblock Bsp-net: Generating compact meshes via binary space partitioning.
\newblock In {\em Proceedings of the IEEE/CVF conference on computer vision and pattern recognition}, pages 45--54, 2020.

\bibitem{chen2024v3d}
Z.~Chen, Y.~Wang, F.~Wang, Z.~Wang, and H.~Liu.
\newblock V3d: Video diffusion models are effective 3d generators.
\newblock {\em arXiv preprint arXiv:2403.06738}, 2024.

\bibitem{cheng2023sdfusion}
Y.-C. Cheng, H.-Y. Lee, S.~Tulyakov, A.~G. Schwing, and L.-Y. Gui.
\newblock Sdfusion: Multimodal 3d shape completion, reconstruction, and generation.
\newblock In {\em Proceedings of the IEEE/CVF Conference on Computer Vision and Pattern Recognition}, pages 4456--4465, 2023.

\bibitem{chou2023diffusion}
G.~Chou, Y.~Bahat, and F.~Heide.
\newblock Diffusion-sdf: Conditional generative modeling of signed distance functions.
\newblock In {\em Proceedings of the IEEE/CVF International Conference on Computer Vision}, pages 2262--2272, 2023.

\bibitem{dai2019scan2mesh}
A.~Dai and M.~Nie{\ss}ner.
\newblock Scan2mesh: From unstructured range scans to 3d meshes.
\newblock In {\em Proceedings of the IEEE/CVF Conference on Computer Vision and Pattern Recognition}, pages 5574--5583, 2019.

\bibitem{deitke2024objaverse}
M.~Deitke, R.~Liu, M.~Wallingford, H.~Ngo, O.~Michel, A.~Kusupati, A.~Fan, C.~Laforte, V.~Voleti, S.~Y. Gadre, et~al.
\newblock Objaverse-xl: A universe of 10m+ 3d objects.
\newblock {\em Advances in Neural Information Processing Systems}, 36, 2024.

\bibitem{deitke2023objaverse}
M.~Deitke, D.~Schwenk, J.~Salvador, L.~Weihs, O.~Michel, E.~VanderBilt, L.~Schmidt, K.~Ehsani, A.~Kembhavi, and A.~Farhadi.
\newblock Objaverse: A universe of annotated 3d objects.
\newblock In {\em Proceedings of the IEEE/CVF Conference on Computer Vision and Pattern Recognition}, pages 13142--13153, 2023.

\bibitem{erkocc2023hyperdiffusion}
Z.~Erko{\c{c}}, F.~Ma, Q.~Shan, M.~Nie{\ss}ner, and A.~Dai.
\newblock Hyperdiffusion: Generating implicit neural fields with weight-space diffusion.
\newblock In {\em Proceedings of the IEEE/CVF International Conference on Computer Vision}, pages 14300--14310, 2023.

\bibitem{groueix2018papier}
T.~Groueix, M.~Fisher, V.~G. Kim, B.~C. Russell, and M.~Aubry.
\newblock A papier-m{\^a}ch{\'e} approach to learning 3d surface generation.
\newblock In {\em Proceedings of the IEEE conference on computer vision and pattern recognition}, pages 216--224, 2018.

\bibitem{gupta3DGenTriplaneLatent2023}
A.~Gupta, W.~Xiong, Y.~Nie, I.~Jones, and B.~O{\u g}uz.
\newblock {{3DGen}}: {{Triplane Latent Diffusion}} for {{Textured Mesh Generation}}, Mar. 2023.

\bibitem{hong2023lrm}
Y.~Hong, K.~Zhang, J.~Gu, S.~Bi, Y.~Zhou, D.~Liu, F.~Liu, K.~Sunkavalli, T.~Bui, and H.~Tan.
\newblock Lrm: Large reconstruction model for single image to 3d.
\newblock {\em arXiv preprint arXiv:2311.04400}, 2023.

\bibitem{hui2022neural}
K.-H. Hui, R.~Li, J.~Hu, and C.-W. Fu.
\newblock Neural wavelet-domain diffusion for 3d shape generation.
\newblock In {\em SIGGRAPH Asia 2022 Conference Papers}, pages 1--9, 2022.

\bibitem{jun2023shap}
H.~Jun and A.~Nichol.
\newblock Shap-e: Generating conditional 3d implicit functions.
\newblock {\em arXiv preprint arXiv:2305.02463}, 2023.

\bibitem{li2023instant3d}
J.~Li, H.~Tan, K.~Zhang, Z.~Xu, F.~Luan, Y.~Xu, Y.~Hong, K.~Sunkavalli, G.~Shakhnarovich, and S.~Bi.
\newblock Instant3d: Fast text-to-3d with sparse-view generation and large reconstruction model.
\newblock {\em arXiv preprint arXiv:2311.06214}, 2023.

\bibitem{linMagic3DHighResolutionTextto3D2023}
C.-H. Lin, J.~Gao, L.~Tang, T.~Takikawa, X.~Zeng, X.~Huang, K.~Kreis, S.~Fidler, M.-Y. Liu, and T.-Y. Lin.
\newblock {{Magic3D}}: {{High-Resolution Text-to-3D Content Creation}}.
\newblock In {\em Proceedings of the {{IEEE}}/{{CVF Conference}} on {{Computer Vision}} and {{Pattern Recognition}}}, 2023.

\bibitem{liuZero1to3ZeroshotOne2023}
R.~Liu, R.~Wu, B.~Van~Hoorick, P.~Tokmakov, S.~Zakharov, and C.~Vondrick.
\newblock Zero-1-to-3: {{Zero-shot One Image}} to {{3D Object}}, Mar. 2023.

\bibitem{liuMeshDiffusionScorebasedGenerative2023}
Z.~Liu, Y.~Feng, M.~J. Black, D.~Nowrouzezahrai, L.~Paull, and W.~Liu.
\newblock {{MeshDiffusion}}: {{Score-based Generative 3D Mesh Modeling}}.
\newblock In {\em The {{Eleventh International Conference}} on {{Learning Representations}}}, Feb. 2023.

\bibitem{lorensen1998marching}
W.~E. Lorensen and H.~E. Cline.
\newblock Marching cubes: A high resolution 3d surface construction algorithm.
\newblock In {\em Seminal graphics: pioneering efforts that shaped the field}, pages 347--353. 1998.

\bibitem{loshchilov2017decoupled}
I.~Loshchilov and F.~Hutter.
\newblock Decoupled weight decay regularization.
\newblock {\em arXiv preprint arXiv:1711.05101}, 2017.

\bibitem{lyu2023controllable}
Z.~Lyu, J.~Wang, Y.~An, Y.~Zhang, D.~Lin, and B.~Dai.
\newblock Controllable mesh generation through sparse latent point diffusion models.
\newblock In {\em Proceedings of the IEEE/CVF conference on computer vision and pattern recognition}, pages 271--280, 2023.

\bibitem{martinez2014stacked}
J.~Martinez, H.~H. Hoos, and J.~J. Little.
\newblock Stacked quantizers for compositional vector compression.
\newblock {\em arXiv preprint arXiv:1411.2173}, 2014.

\bibitem{muller2023diffrf}
N.~M{\"u}ller, Y.~Siddiqui, L.~Porzi, S.~R. Bulo, P.~Kontschieder, and M.~Nie{\ss}ner.
\newblock Diffrf: Rendering-guided 3d radiance field diffusion.
\newblock In {\em Proceedings of the IEEE/CVF Conference on Computer Vision and Pattern Recognition}, pages 4328--4338, 2023.

\bibitem{nash2020polygen}
C.~Nash, Y.~Ganin, S.~A. Eslami, and P.~Battaglia.
\newblock Polygen: An autoregressive generative model of 3d meshes.
\newblock In {\em International conference on machine learning}, pages 7220--7229. PMLR, 2020.

\bibitem{pooleDreamFusionTextto3DUsing2023}
B.~Poole, A.~Jain, J.~T. Barron, and B.~Mildenhall.
\newblock {{DreamFusion}}: {{Text-to-3D}} using {{2D Diffusion}}.
\newblock In {\em {{ICLR}}}. {arXiv}, 2023.

\bibitem{rombachHighResolutionImageSynthesis2022}
R.~Rombach, A.~Blattmann, D.~Lorenz, P.~Esser, and B.~Ommer.
\newblock High-{{Resolution Image Synthesis With Latent Diffusion Models}}.
\newblock In {\em Proceedings of the {{IEEE}}/{{CVF Conference}} on {{Computer Vision}} and {{Pattern Recognition}}}, pages 10684--10695, 2022.

\bibitem{saharia2022photorealistic}
C.~Saharia, W.~Chan, S.~Saxena, L.~Li, J.~Whang, E.~Denton, S.~K.~S. Ghasemipour, B.~K. Ayan, S.~S. Mahdavi, R.~G. Lopes, T.~Salimans, J.~Ho, D.~J. Fleet, and M.~Norouzi.
\newblock Photorealistic text-to-image diffusion models with deep language understanding, 2022.

\bibitem{shen2021deep}
T.~Shen, J.~Gao, K.~Yin, M.-Y. Liu, and S.~Fidler.
\newblock Deep marching tetrahedra: a hybrid representation for high-resolution 3d shape synthesis.
\newblock {\em Advances in Neural Information Processing Systems}, 34:6087--6101, 2021.

\bibitem{shi2023zero123++}
R.~Shi, H.~Chen, Z.~Zhang, M.~Liu, C.~Xu, X.~Wei, L.~Chen, C.~Zeng, and H.~Su.
\newblock Zero123++: a single image to consistent multi-view diffusion base model.
\newblock {\em arXiv preprint arXiv:2310.15110}, 2023.

\bibitem{shi2023mvdream}
Y.~Shi, P.~Wang, J.~Ye, M.~Long, K.~Li, and X.~Yang.
\newblock Mvdream: Multi-view diffusion for 3d generation.
\newblock {\em arXiv preprint arXiv:2308.16512}, 2023.

\bibitem{shim2023diffusion}
J.~Shim, C.~Kang, and K.~Joo.
\newblock Diffusion-based signed distance fields for 3d shape generation.
\newblock In {\em Proceedings of the IEEE/CVF Conference on Computer Vision and Pattern Recognition}, pages 20887--20897, 2023.

\bibitem{siddiqui2023meshgpt}
Y.~Siddiqui, A.~Alliegro, A.~Artemov, T.~Tommasi, D.~Sirigatti, V.~Rosov, A.~Dai, and M.~Nie{\ss}ner.
\newblock Meshgpt: Generating triangle meshes with decoder-only transformers.
\newblock {\em arXiv preprint arXiv:2311.15475}, 2023.

\bibitem{tang2024lgm}
J.~Tang, Z.~Chen, X.~Chen, T.~Wang, G.~Zeng, and Z.~Liu.
\newblock Lgm: Large multi-view gaussian model for high-resolution 3d content creation.
\newblock {\em arXiv preprint arXiv:2402.05054}, 2024.

\bibitem{Touvron2023Llama2O}
H.~Touvron, L.~Martin, K.~R. Stone, P.~Albert, A.~Almahairi, Y.~Babaei, N.~Bashlykov, S.~Batra, P.~Bhargava, S.~Bhosale, D.~M. Bikel, L.~Blecher, C.~C. Ferrer, M.~Chen, G.~Cucurull, D.~Esiobu, J.~Fernandes, J.~Fu, W.~Fu, B.~Fuller, C.~Gao, V.~Goswami, N.~Goyal, A.~S. Hartshorn, S.~Hosseini, R.~Hou, H.~Inan, M.~Kardas, V.~Kerkez, M.~Khabsa, I.~M. Kloumann, A.~V. Korenev, P.~S. Koura, M.-A. Lachaux, T.~Lavril, J.~Lee, D.~Liskovich, Y.~Lu, Y.~Mao, X.~Martinet, T.~Mihaylov, P.~Mishra, I.~Molybog, Y.~Nie, A.~Poulton, J.~Reizenstein, R.~Rungta, K.~Saladi, A.~Schelten, R.~Silva, E.~M. Smith, R.~Subramanian, X.~Tan, B.~Tang, R.~Taylor, A.~Williams, J.~X. Kuan, P.~Xu, Z.~Yan, I.~Zarov, Y.~Zhang, A.~Fan, M.~Kambadur, S.~Narang, A.~Rodriguez, R.~Stojnic, S.~Edunov, and T.~Scialom.
\newblock Llama 2: Open foundation and fine-tuned chat models.
\newblock {\em ArXiv}, abs/2307.09288, 2023.

\bibitem{voleti2024sv3d}
V.~Voleti, C.-H. Yao, M.~Boss, A.~Letts, D.~Pankratz, D.~Tochilkin, C.~Laforte, R.~Rombach, and V.~Jampani.
\newblock Sv3d: Novel multi-view synthesis and 3d generation from a single image using latent video diffusion.
\newblock {\em arXiv preprint arXiv:2403.12008}, 2024.

\bibitem{wang2018pixel2mesh}
N.~Wang, Y.~Zhang, Z.~Li, Y.~Fu, W.~Liu, and Y.-G. Jiang.
\newblock Pixel2mesh: Generating 3d mesh models from single rgb images.
\newblock In {\em Proceedings of the European conference on computer vision (ECCV)}, pages 52--67, 2018.

\bibitem{wang2023rodin}
T.~Wang, B.~Zhang, T.~Zhang, S.~Gu, J.~Bao, T.~Baltrusaitis, J.~Shen, D.~Chen, F.~Wen, Q.~Chen, et~al.
\newblock Rodin: A generative model for sculpting 3d digital avatars using diffusion.
\newblock In {\em Proceedings of the IEEE/CVF conference on computer vision and pattern recognition}, pages 4563--4573, 2023.

\bibitem{wangProlificDreamerHighFidelityDiverse2023}
Z.~Wang, C.~Lu, Y.~Wang, F.~Bao, C.~Li, H.~Su, and J.~Zhu.
\newblock {{ProlificDreamer}}: {{High-Fidelity}} and {{Diverse Text-to-3D Generation}} with {{Variational Score Distillation}}, May 2023.

\bibitem{wang2024crm}
Z.~Wang, Y.~Wang, Y.~Chen, C.~Xiang, S.~Chen, D.~Yu, C.~Li, H.~Su, and J.~Zhu.
\newblock Crm: Single image to 3d textured mesh with convolutional reconstruction model.
\newblock {\em arXiv preprint arXiv:2403.05034}, 2024.

\bibitem{weng2023consistent123}
H.~Weng, T.~Yang, J.~Wang, Y.~Li, T.~Zhang, C.~Chen, and L.~Zhang.
\newblock Consistent123: Improve consistency for one image to 3d object synthesis.
\newblock {\em arXiv preprint arXiv:2310.08092}, 2023.

\bibitem{xu2024instantmesh}
J.~Xu, W.~Cheng, Y.~Gao, X.~Wang, S.~Gao, and Y.~Shan.
\newblock Instantmesh: Efficient 3d mesh generation from a single image with sparse-view large reconstruction models.
\newblock {\em arXiv preprint arXiv:2404.07191}, 2024.

\bibitem{xu2024grm}
Y.~Xu, Z.~Shi, W.~Yifan, H.~Chen, C.~Yang, S.~Peng, Y.~Shen, and G.~Wetzstein.
\newblock Grm: Large gaussian reconstruction model for efficient 3d reconstruction and generation.
\newblock {\em arXiv preprint arXiv:2403.14621}, 2024.

\bibitem{xu2023dmv3d}
Y.~Xu, H.~Tan, F.~Luan, S.~Bi, P.~Wang, J.~Li, Z.~Shi, K.~Sunkavalli, G.~Wetzstein, Z.~Xu, et~al.
\newblock Dmv3d: Denoising multi-view diffusion using 3d large reconstruction model.
\newblock {\em arXiv preprint arXiv:2311.09217}, 2023.

\bibitem{zhang20233dshape2vecset}
B.~Zhang, J.~Tang, M.~Niessner, and P.~Wonka.
\newblock 3dshape2vecset: A 3d shape representation for neural fields and generative diffusion models.
\newblock {\em ACM Transactions on Graphics (TOG)}, 42(4):1--16, 2023.

\bibitem{zheng2023free3d}
C.~Zheng and A.~Vedaldi.
\newblock Free3d: Consistent novel view synthesis without 3d representation.
\newblock {\em arXiv preprint arXiv:2312.04551}, 2023.

\bibitem{zheng2023locally}
X.-Y. Zheng, H.~Pan, P.-S. Wang, X.~Tong, Y.~Liu, and H.-Y. Shum.
\newblock Locally attentional sdf diffusion for controllable 3d shape generation.
\newblock {\em ACM Transactions on Graphics (TOG)}, 42(4):1--13, 2023.

\end{thebibliography}
}


\clearpage

\appendix

\section{Additional Results}

\subsection{Comparison with InstantMesh}

\begin{figure}[ht]
\centering
\includegraphics[width=\textwidth]{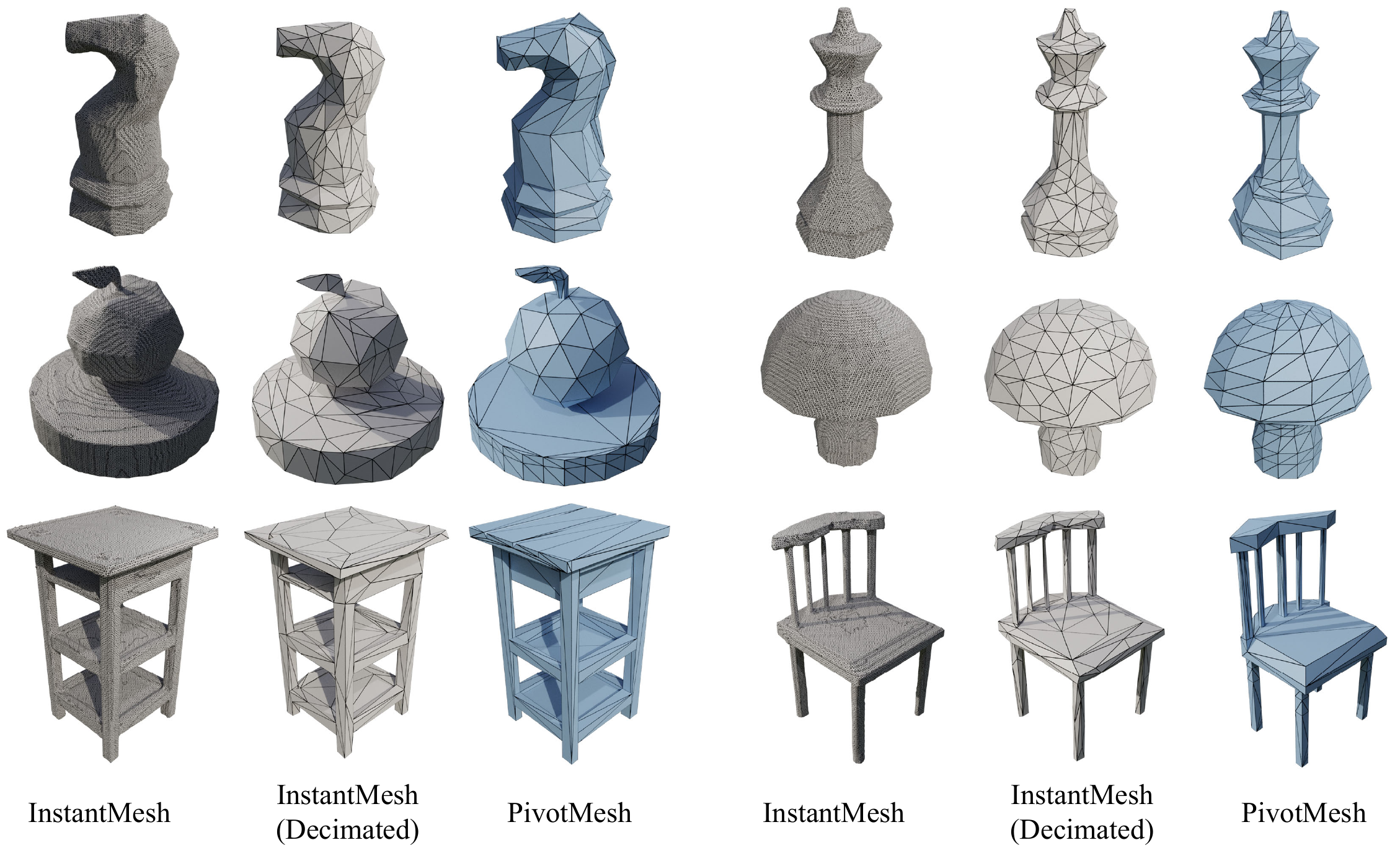}
\caption{\textbf{Additional results for comparison with InstantMesh.} 
}
\label{fig: suppl-instantmesh}
\end{figure}

To produce similar shapes for a more straightforward visual comparison, we use the rendering images of our generated mesh as the image condition for InstantMesh to generate the following instances in Figure \ref{fig: head} and Figure \ref{fig: suppl-instantmesh}.

\subsection{Unconditional Generation on ShapeNet}

\begin{figure}[ht]
\centering
\includegraphics[width=\textwidth]{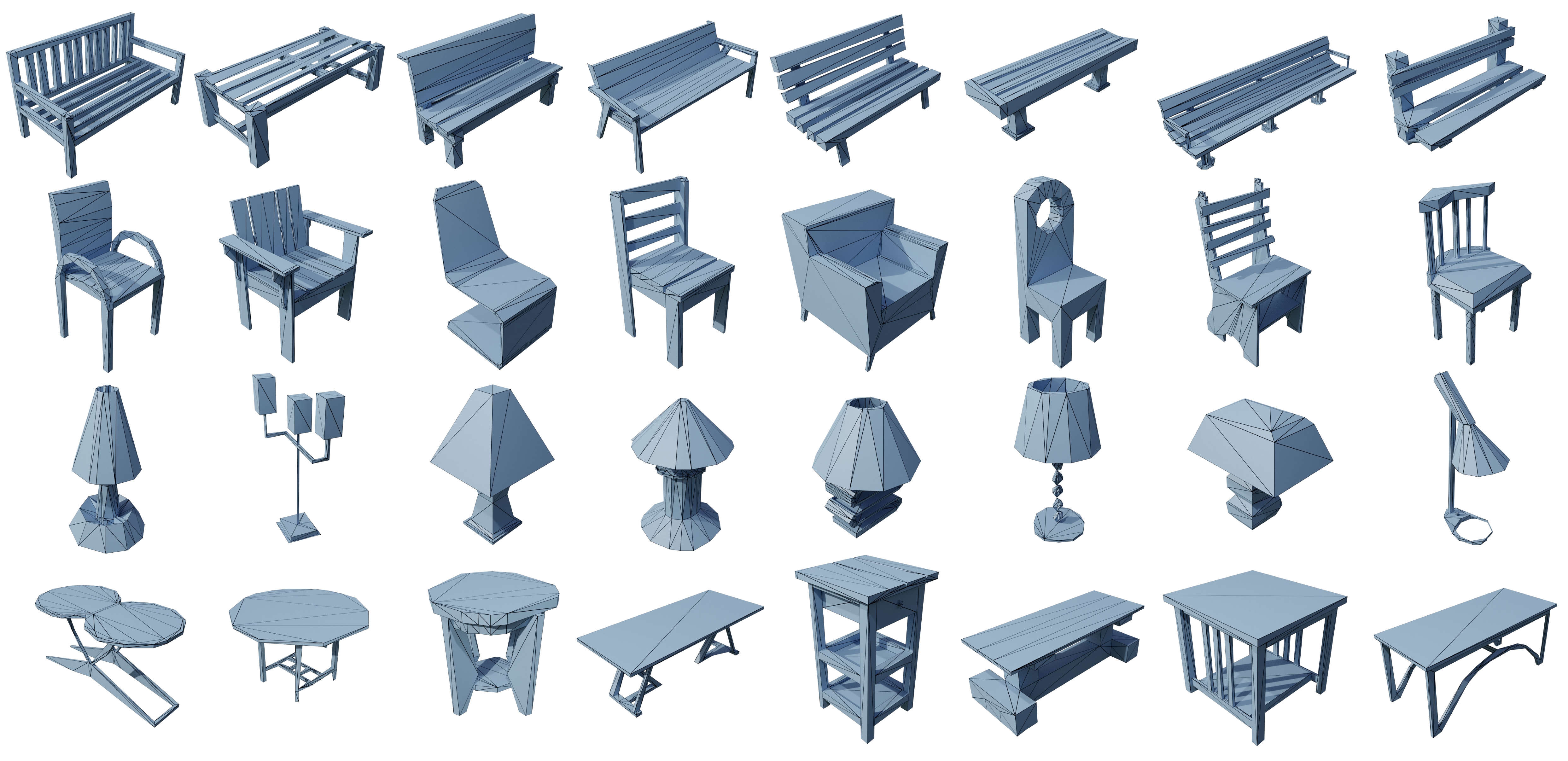}
\caption{\textbf{Additional results for unconditional generation on Shapenet.} 
}
\label{fig: suppl-shapenet}
\end{figure}

\clearpage

\subsection{Unconditional Generation on Objaverse}

\begin{figure}[ht]
\centering
\includegraphics[width=\textwidth]{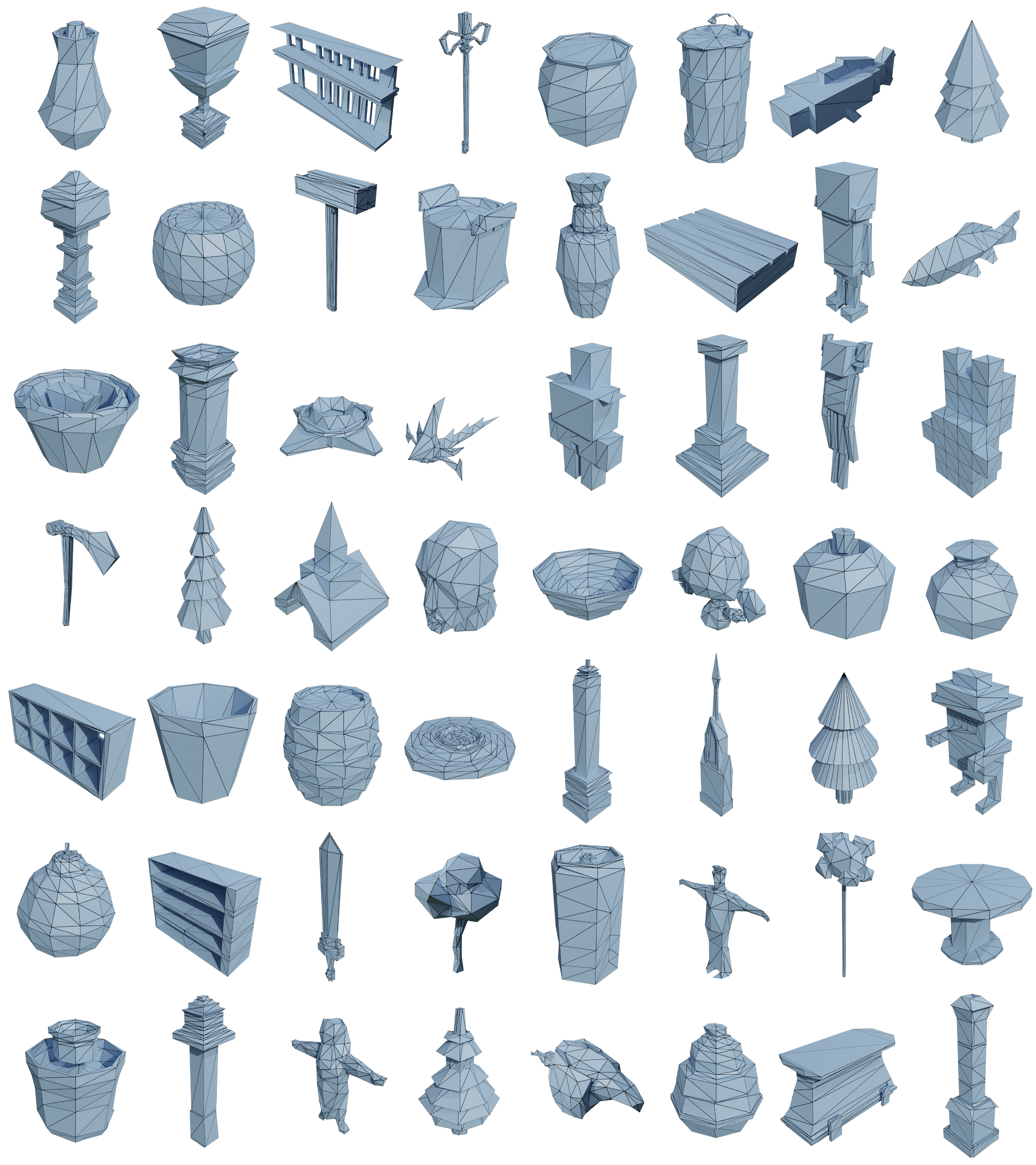}
\caption{\textbf{Additional results for unconditional generation on Objaverse.} 
}
\label{fig: suppl-objaverse}
\end{figure}

\clearpage

\subsection{Pivot-guided Mesh Generation}

\begin{figure}[ht]
\centering
\includegraphics[width=\textwidth]{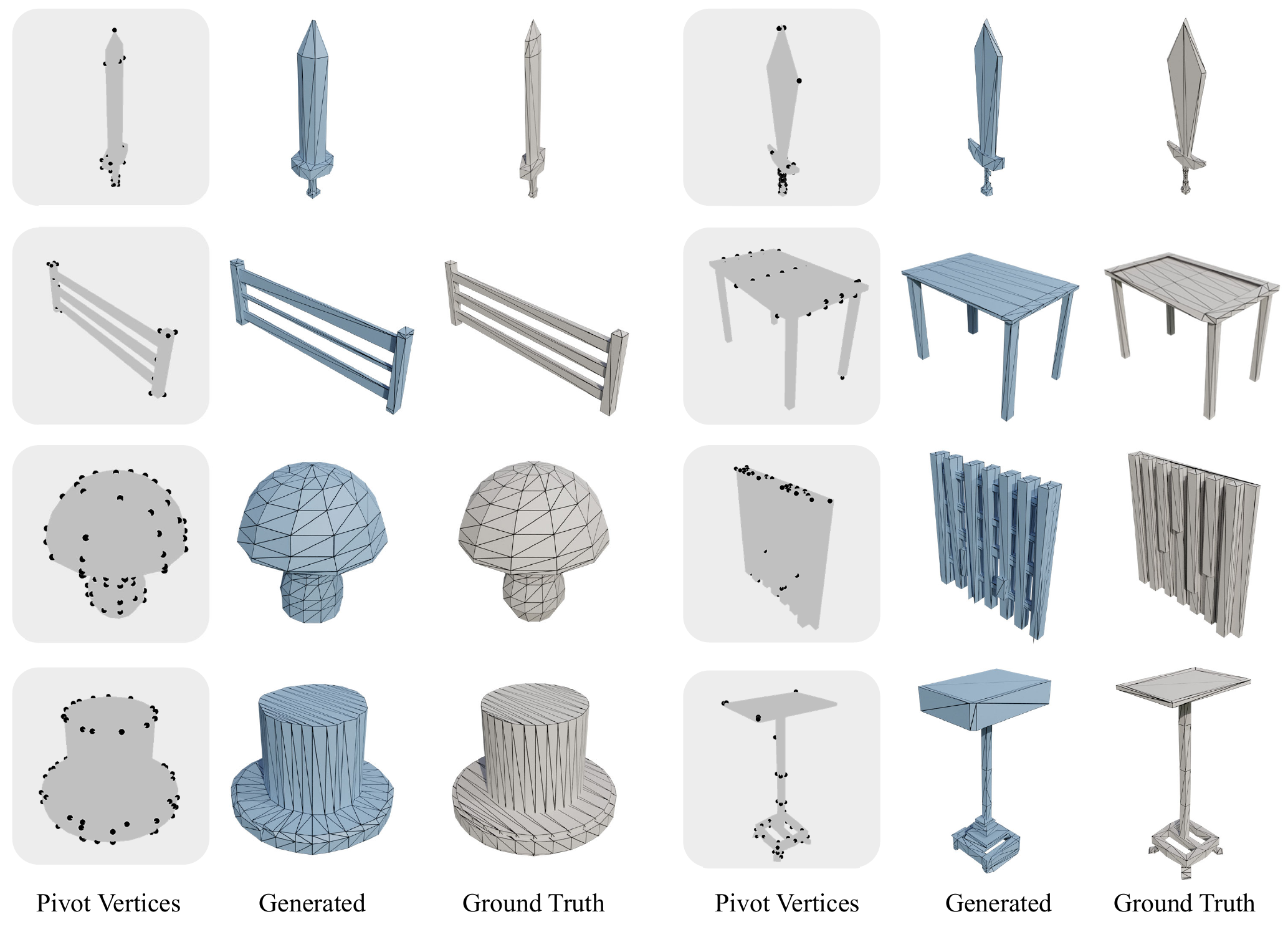}
\caption{\textbf{Additional results for pivot-guided mesh generation on Objaverse.} 
}
\label{fig: suppl-cond}
\end{figure}

\subsection{Shape Novelty Analysis}

\begin{figure}[ht]
\centering
\includegraphics[width=\textwidth]{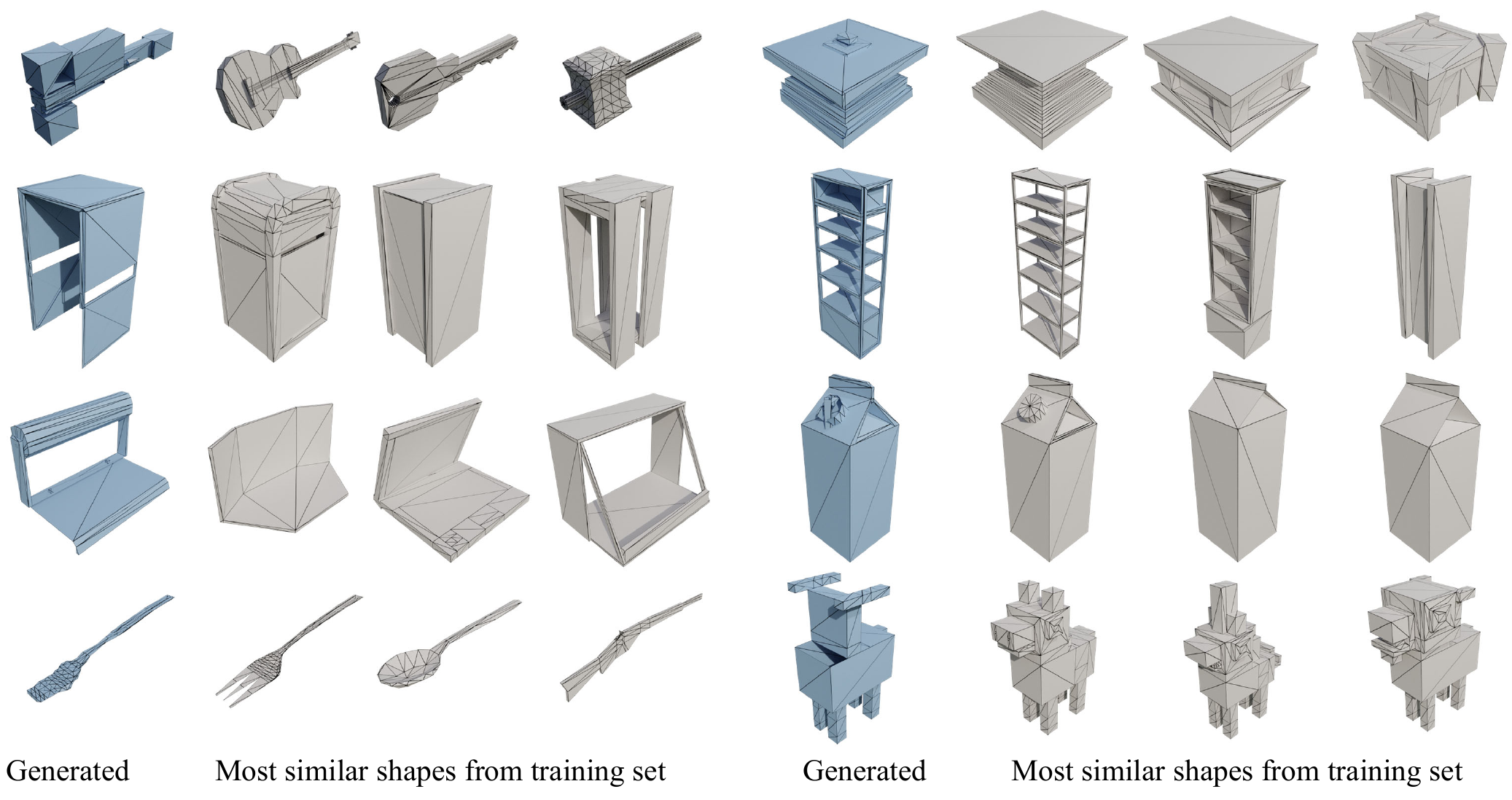}
\caption{\textbf{Additional results for shape novelty analysis on Objaverse.} 
}
\label{fig: suppl-novel}
\end{figure}

\clearpage

\subsection{PivotMesh Applications}

\begin{figure}[ht]
\centering
\includegraphics[width=\textwidth]{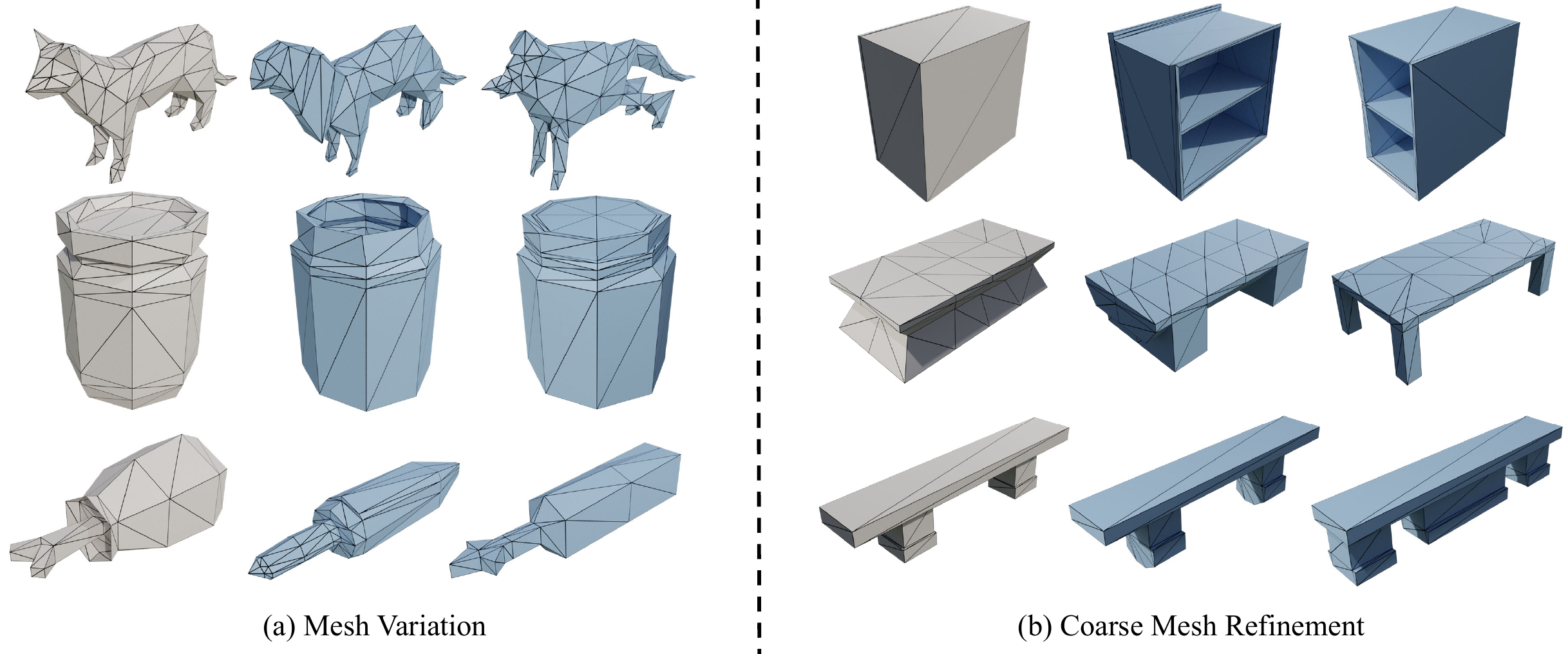}
\caption{\textbf{Additional results for the applications of PivotMesh.} 
}
\label{fig: suppl-applications}
\end{figure}

\end{document}